\documentclass[twoside]{article}

\usepackage[accepted]{aistats2026}
\usepackage[round]{natbib}

\usepackage{graphicx}
\usepackage{hyperref}
\usepackage{pifont}
\usepackage{algorithm}
\usepackage{algpseudocode}

\begin{document}

%

%

\twocolumn[

\aistatstitle{Three-Step Nav: A Hierarchical Global–Local Planner for Zero-Shot Vision-and-Language Navigation}

\aistatsauthor{ Wanrong Zheng$^1$ \And Yunhao Ge$^2$ \And Laurent Itti$^1$ }

\aistatsaddress{ $^1$University of Southern California \And $^2$NVIDIA Research } ]

\begin{abstract}
  
    Breakthrough progress in vision-based navigation through unknown environments has been achieved by using multimodal large language models (MLLMs). These models can plan a sequence of motions by evaluating the current view at each time step against the task and goal given to the agent. However, current zero-shot Vision-and-Language Navigation (VLN) agents powered by MLLMs still tend to drift off course, halt prematurely, and achieve low overall success rates. We propose Three-Step Nav to counteract these failures with a three-view protocol: First, "look forward" to extract global landmarks and sketch a coarse plan. Then, "look now" to align the current visual observation with the next sub-goal for fine-grained guidance. Finally, "look backward" audits the entire trajectory to correct accumulated drift before stopping. Requiring no gradient updates or task-specific fine-tuning, our planner drops into existing VLN pipelines with minimal overhead. Three-Step Nav achieves state-of-the-art zero-shot performance on the R2R-CE and RxR-CE dataset. Our code is available at \url{https://github.com/ZoeyZheng0/3-step-Nav}.
\end{abstract}

\begin{figure}[!t] 
    \centering
    \includegraphics[width=\columnwidth]{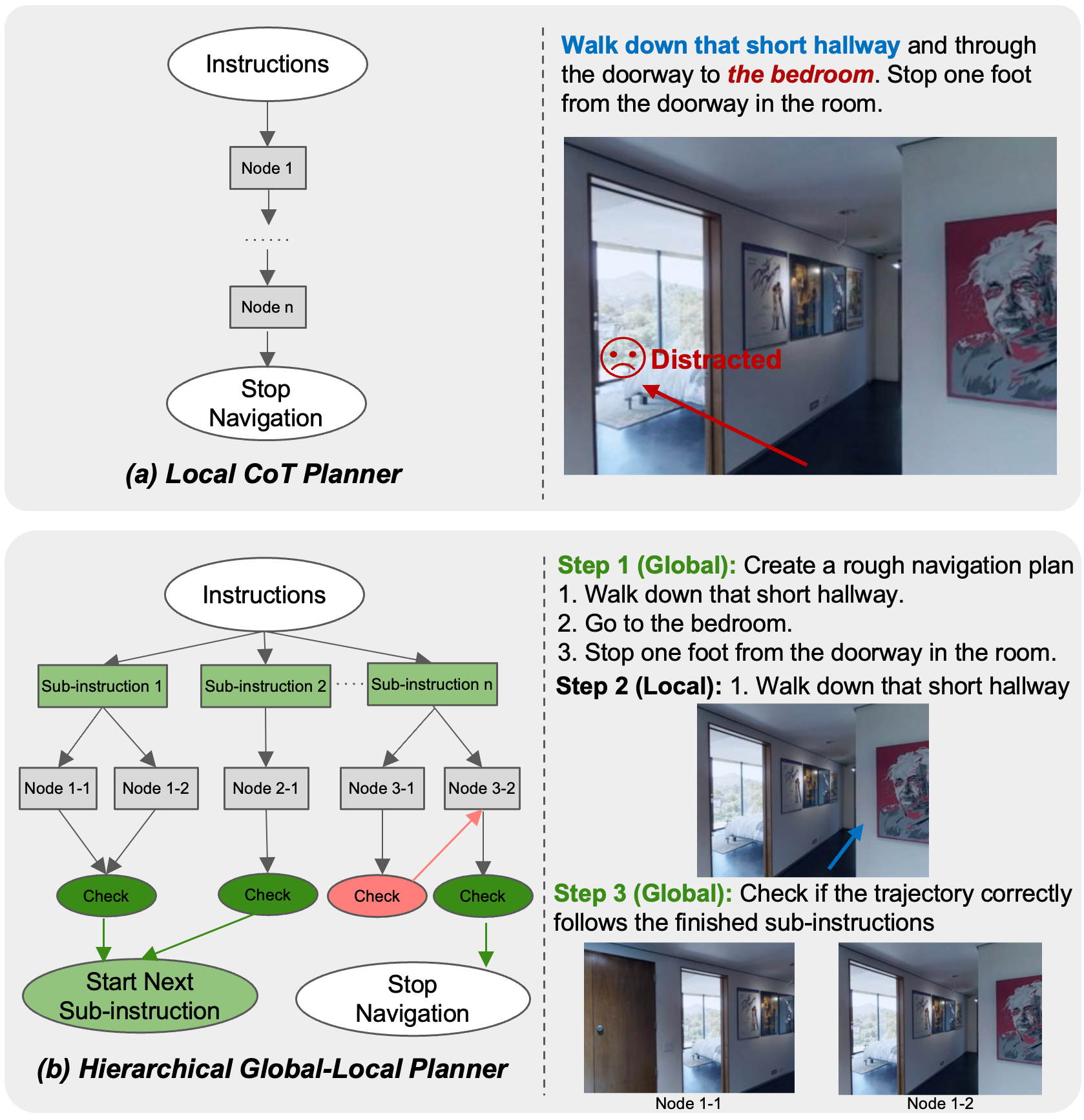}
    \caption{ (a) Prior LLM-core planners rely only on the current RGB-D view and textual action history, often misjudging progress and getting distracted by irrelevant objects. In contrast, (b) our Hierarchical Global–Local Planner mitigates this by first decomposing instructions into sub-instructions for a global plan, then locally grounding each sub-instruction, and finally verifying completed sub-instructions against the global trajectory—yielding more robust zero-shot navigation. }
    \label{fig:motivation}
\end{figure}

\section{INTRODUCTION}

In the Vision-and-Language Navigation (VLN) \citep{anderson2018vision, gu2022vision, wang2023scaling, li2024vln} task, embodied agents are required to navigate to an unseen destination following a series of natural language instructions. Early studies \citep{an2024etpnav, zhou2024navgpt2} simplified the task by grounding navigation in discrete graphs, where the agent selects between predefined viewpoints and edges depending on RGB-D sensory inputs. Although effective for benchmarking, this abstraction neglects the low-level dynamics of real-world robotics, such as continuous motion, partial observability, and collision risks. To bridge this gap, the community has introduced VLN in Continuous Environments (VLN-CE) \citep{krantz2020beyond}, which eliminates the dependence on connectivity graphs and instead equips agents with egocentric sensors and low-level actions. This setting is substantially more challenging than discrete graph-based VLN, as it requires reasoning over open spaces, handling unforeseen obstacles, and maintaining progress without predefined navigation graphs. Building agents that can solve VLN-CE reliably represents a critical step toward general-purpose embodied AI, enabling real-world applications such as home robotics, assistive navigation, and search and rescue operations.

Meanwhile, Multimodal Large Language Models (MLLMs) have demonstrated strong zero-shot abilities in vision-language tasks \citep{zhou2024navgpt, long2024discuss, chen2024mapgpt}. Their broad world knowledge and flexible reasoning suggest a new path for VLN: treat navigation as an iterative decision-making dialogue with a powerful but generic model, rather than training a bespoke policy from scratch.  However, directly requiring an MLLM faces two key challenges in continuous VLN: As shown in Fig.~\ref{fig:motivation} (a), instructions can span dozens of steps; the agent must balance global route planning with local actuation, yet commodity MLLMs reason over a limited context window. Moreover, in continuous space, small heading or position errors quickly compound. The agent needs principled mechanisms to detect mistakes and recover without explicit supervision.

To address these issues, we present \emph{Three-Step Nav}, a hierarchical global–local framework that leverages MLLMs for zero-shot VLN in continuous 3-D environments, differing from prior MLLM-based VLN agents by coupling global planning with trajectory-level verification. The agent alternates a global--local--global reasoning loop: (i) \emph{looking forward} to outline upcoming sub-instructions, (ii) \emph{looking now} to ground the current sub-goal in live visual observations, and (iii) \emph{looking back} to verify past progress and adjust future plans.  Within this loop, we introduce an \emph{adaptive judge module} that endows the agent with four meta-skills - \emph{stay}, \emph{continue}, \emph{backtrack}, and \emph{look-around} - allowing dynamic self-correction when uncertainty is detected.

Compared with prior LLM-core planners, our hierarchical global–local design achieves state-of-the-art zero-shot success rates R2R-CE \citep{ku2020room} and RxR-CE \citep{krantz2020beyond} datasets, while also reducing navigation error by 15\% and improving SPL by 12\% on the validation-unseen splits of R2R-CE, indicating that our global progress check and trajectory-level auditing effectively mitigate distraction and cumulative drift in continuous environments. The goals of this work can be summarized as follows:
\begin{itemize}
    \item We propose Three-Step Nav, a novel framework that alternates global–local–global reasoning with an MLLM, enabling zero-shot VLN in continuous 3-D environments while preserving long-range context and requiring no task-specific fine-tuning. Moreover, the framework is lightweight and modular, making it easy to plug into other LLM-core planning pipelines.

    \item We introduce a Hierarchical Global–Local Planner that dynamically switches views: after completing local chain-of-thought reasoning to reach a specific sub-goal, the agent transitions to a global check that examines the trajectory and verifies finished sub-instructions. This alternating loop between fine-grained execution and trajectory-level auditing helps the MLLM reason over spatial structure and exploration history, suppressing distraction and cumulative drift.

    \item We design an adaptive judge module equipped with meta-skills to decide whether to \emph{stay}, \emph{continue}, \emph{backtrack}, or \emph{look around}, allowing the agent to self-correct under uncertainty and maintain robust navigation even in ambiguous environments.

\end{itemize}

\begin{figure*}[!t] 
    \centering
    \includegraphics[width=\textwidth]{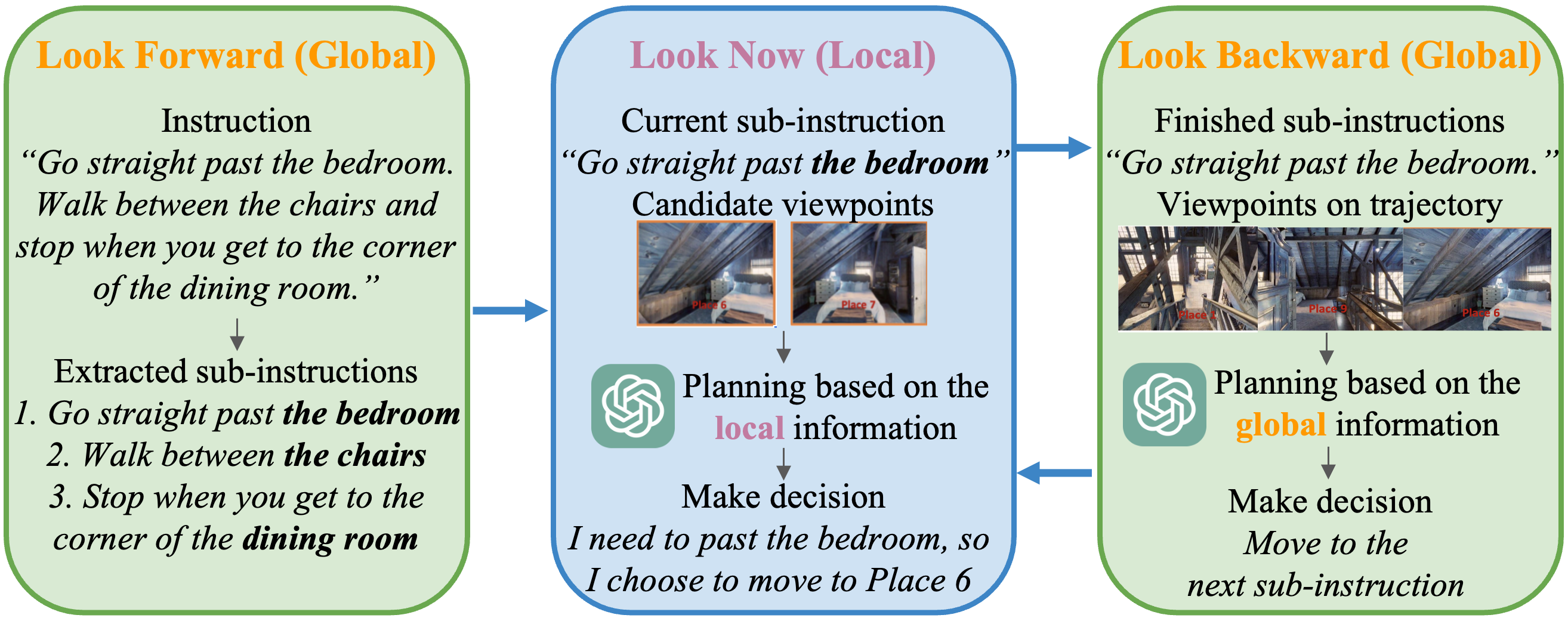}
    \caption{Illustration of the overall pipeline of the proposed methodology. We have three modules: \emph{look forward}, \emph{look now}, and \emph{look backward}. The agent first looks forward to decomposing the natural-language instruction into an ordered list of sub-instructions and to extracting salient global landmarks that sketch a coarse route. 
    Next, it looks now by matching the current observation against the active sub-instruction, and selecting the next waypoint for fine-grained local progress. 
    Finally, it looks backward to audit the trajectory completed so far—revisiting stored viewpoints, verifying that finished sub-instructions were indeed satisfied, and triggering corrective backtracking if drift is detected. }
    \label{fig:overall}
\end{figure*}

\section{RELATED WORK}
\paragraph{Vision-and-Language Navigation}
The VLN task \citep{anderson2018vision} requires embodied agents to follow natural language instructions and visual observations to reach goal locations in novel environments. Early research in the discrete VLN setting, typically built on the Matterport3D simulator \citep{chang2017matterport3d}, modeled navigation as sequential decision making on a predefined connectivity graph. To enhance performance, methods introduced various structural priors: DUET \citep{chen2022think} and ETPNav \citep{an2024etpnav} leveraged topological abstractions to capture global spatial relations, BEVBert \citep{an2023bevbert} constructed semantic bird’s-eye view representations, and \citet{wang2023gridmm} designed egocentric grid-based memory for long-horizon reasoning. Beyond graph-based or memory-driven methods, VLN-Video \citep{li2024vln} utilized driving videos to extend VLN into outdoor navigation, and \citet{zhu2023vision} introduced knowledge-driven imagination of unseen layouts. Collectively, these approaches advanced discrete VLN by enriching agents’ spatial reasoning and grounding, laying the foundation for more realistic navigation formulations. Different from these training-intensive approaches, our work explores training-free, zero-shot VLN framework, which employing MLLMs as core planners.

\paragraph{Navigation with MLLM}
LLMs have demonstrated remarkable generalization and reasoning capabilities, sparking significant interest in their application to navigation tasks. \citet{zhou2024navgpt} introduced NavGPT, a purely LLM-based navigation agent that performs zero-shot sequential action prediction in VLN tasks by utilizing textual descriptions of visual observations, navigation history, and future explorable directions. Building upon this, a subsequent work \citet{zhou2024navgpt2} aimed to bridge the gap between LLM-based agents and VLN-specialized models by aligning visual content within a frozen LLM and incorporating navigation policy networks. Navid \citep{zhang2024navid} adapted Vicuna-based LLMs to embodied navigation, while MapGPT \citep{chen2024mapgpt} and DiscussNav \citep{long2024discuss} explored topological textual map-guided exploration and multi-expert collaboration. Furthermore, recent studies have emphasized that token reduction in generative models should move beyond mere efficiency to improve representation quality across vision and language tasks \citep{kong2025}. These prior studies have largely adopted a step-by-step navigation paradigm that leverages textual memory to maintain long-term context, but such designs are prone to accumulated drift and may suffer from inaccuracies in progress estimation. In contrast, our method introduces explicit global–local reasoning with trajectory-level verification, providing a more structured understanding of navigation.  

\paragraph{MLLM-Core Planner in VLN-CE}
Early work on VLN assumed discrete graph navigation, but the VLN-CE formulation of \citet{krantz2020beyond} exposed the far tougher problem of planning in photorealistic continuous spaces, where agents must issue velocity commands while coping with long horizons, compounding pose error, and partial observability. Open-Nav \citep{qiao2025open} addressed this by coupling constraint reasoning with backtracking to enhance zero-shot robustness, while CA-Nav \citep{chen2025canav} cast zero-shot VLN-CE as sequential constraint satisfaction to ensure both language and spatial criteria are met. More recently, SmartWay \citep{shi2025smartwayenhancedwaypointprediction} integrated waypoint prediction with trajectory-level consistency checks. Different from these approaches, our framework not only enables MLLMs to conduct semantic analysis of the current environment for local navigation among candidate directions, but also equips them with meta-abilities to support higher-level decision making and a deeper understanding of the navigation process.

\begin{figure*}[!t] 
    \centering
    \includegraphics[width=\textwidth]{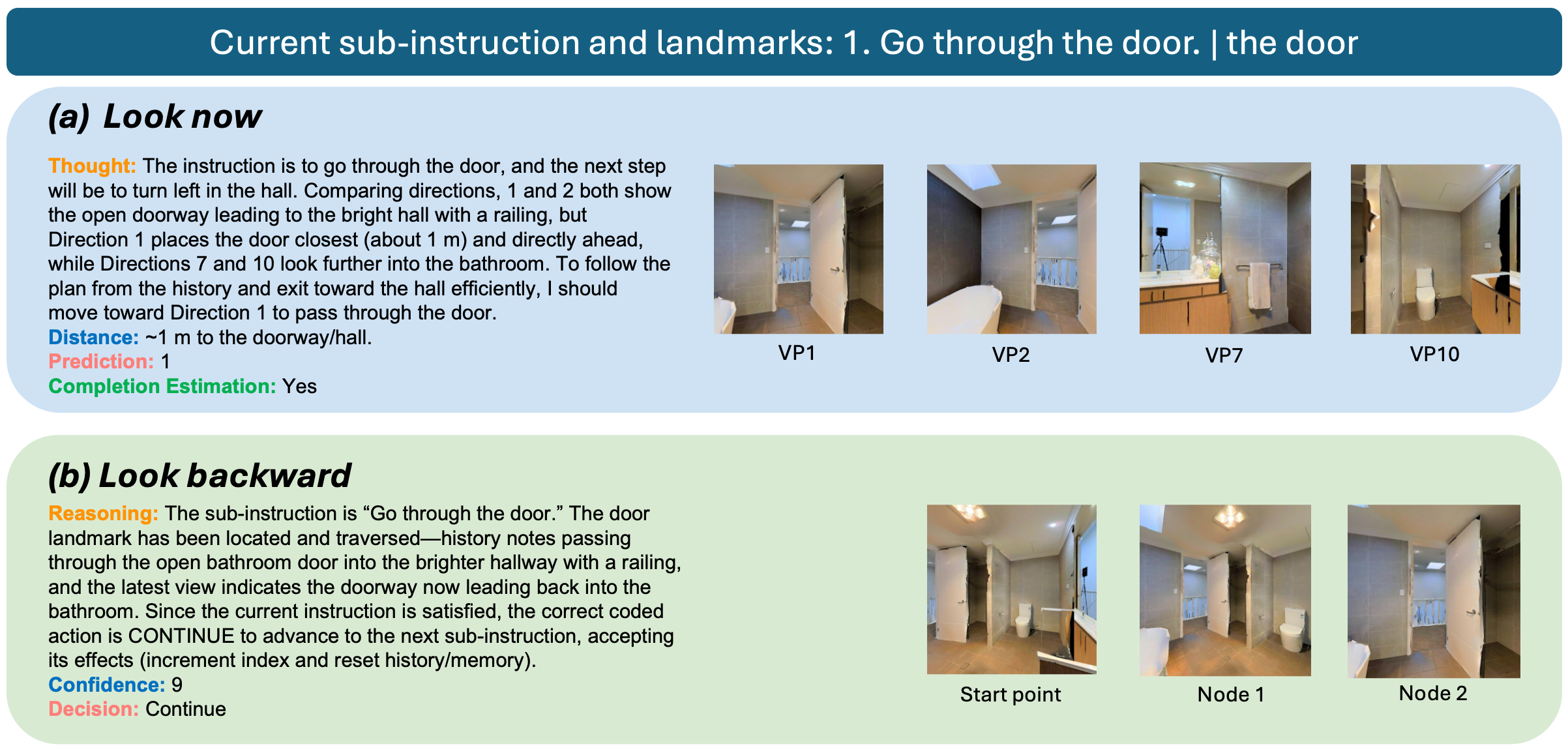}
    \caption{One successful example in the R2R-CE dataset.
    (a) \emph{Look now}. Display the current candidate viewpoints and their short descriptions; the MLLM selects the option most aligned with the active sub-goal $s_k$ as the next direction $v_t^\star$.
    (b) \emph{Look backward}. Visualize all previously chosen viewpoints to form the executed trajectory $\tau$; the MLLM audits $\tau$ against the finished sub-instructions to decide whether the completed sub-goals is satisfied.}
    \label{fig:example}
\end{figure*}

\section{METHOD}
\subsection{Problem Formulation}
Vision-and-Language Navigation in Continuous Environments (VLN-CE) is defined by an autonomous agent navigating within a continuous 3D environment $\mathcal{E}$ to reach a specified goal location based on natural language instructions. Let $x_t$ denote the agent’s pose (position and orientation) at time $t$. At each time step, the agent receives an observation $o_t \in \mathcal{O}$ consisting of a panoramic RGB-D view from its current pose (e.g., 12 RGB images $I_{t,1}^{\mathrm{RGB}}, \dots, I_{t,12}^{\mathrm{RGB}}$ and 12 depth images $I_{t,1}^{\mathrm{D}}, \dots, I_{t,12}^{\mathrm{D}}$ captured at 30° intervals for a full $360^\circ$ panorama). The agent is also given a natural language instruction $W = (w_1, w_2, \ldots, w_L)$, where each $w_i$ is a word token and $L$ is the instruction length, describing the instructions to reach the goal for the agent. The action space $\mathcal{A}$ comprises discrete low-level navigation actions (e.g., turning or moving forward by fixed increments) that change the agent’s pose. Starting from an initial pose $x_0$, the agent iteratively selects an action $a_t \in \mathcal{A}$ based on the instruction $W$ and current observation $o_t$, yielding a new state $x_{t+1}$ and observation $o_{t+1}$ at the next step. This perception–action loop repeats until the agent executes a stop action upon reaching the destination. By following the instruction in this sequential decision process, the agent aims to minimize navigation errors and to successfully arrive at the target location in $\mathcal{E}$.

\subsection{Overview}
As shown in Fig.~\ref{fig:overall}, our methodology consists of a lightweight three-view planner, dubbed Three-Step Nav, that augments any frozen multimodal LLM through prompt engineering. Previous Local Chain-of-Thought (CoT) approaches, which make decisions solely based on the current observation, are limited to reasoning over candidate viewpoints and a textual history memory. As a result, they are prone to drifting or becoming distracted over long trajectories. To address this issue, we introduce a \textbf{Hierarchical Global–Local Planner} that organizes navigation into three complementary stages: \emph{look forward}, \emph{look now}, and \emph{look backward}. In the \emph{look forward} stage, the instruction is decomposed into an ordered sequence of sub-instructions, each anchored by salient global landmarks, which together form a coarse-grained plan. Navigation then proceeds in a loop over these sub-goals. For each sub-goal, the agent repeatedly performs \emph{look now}, selecting from candidate viewpoints the option most aligned with the current sub-instruction and its associated landmark. This local decision-making is guided by the MLLM navigator, which also checks whether the agent has approached the target landmark closely enough to consider the sub-goal completed. Once a sub-goal is tentatively reached, the agent executes \emph{look backward}, auditing its past trajectory by replaying visited viewpoints and verifying that completed sub-instructions and landmarks were indeed satisfied from a global perspective. Only if the MLLM navigator confirms consistency with the history does the planner advance to the next sub-goal; otherwise, it triggers backtracking or refinement. This hierarchical global–local structure enables the agent to maintain long-horizon coherence while still making fine-grained, adaptive local decisions, effectively reducing premature stops and accumulated drift in zero-shot navigation. The methods are summarized in sections \ref{looking_forward}, \ref{looking_now}, and \ref{looking_backward} below.

\subsection{Look forward}
\label{looking_forward}
The looking-forward module converts the full natural-language instructions into an ordered “road-map’’ of sub-instructions and coarse global waypoints before any physical movement begins. Prompted with the instruction text and the agent’s initial panoramic view, the frozen multimodal LLM (i) segments the instruction wherever spatial connectives (e.g., “past,” “between,” “until”) or punctuation mark distinct goals, producing a sequence of atomic sub-instructions, and (ii) highlights the salient nouns that refer to persistent landmarks—rooms, furniture, or objects—that can anchor long-range navigation. Each sub-instruction is then paired with its corresponding landmark, yielding a high-level plan that sketches a straight-line route through the environment. Because this step is purely prompt-based and training-free, it can be injected into any VLN pipeline with negligible overhead while providing the global context that subsequent modules exploit for local decision making and drift correction.

\subsection{Look now}
\label{looking_now}
The looking-now module (Fig. ~\ref{fig:example} (a)) is the agent’s fine-grained decision engine at each time step. Given the current sub-instruction $s_k$ (provided by the look-forward module) and the agent’s observation $o_t$ at pose $x_t$, the agent must decide where to move next. We first enumerate a set of navigable candidate viewpoints $\mathcal{V}t = {v_{t,1}, v_{t,2}, \dots, v_{t,m}}$ reachable from $x_t$ (e.g. adjacent viewpoints returned by the simulator).  Followed by prior work \citep{qiao2025open}, we use a transformer-based model to serve as the Waypoint Prediction module and take panoramic RGB and depth images to pinpoint potential navigation waypoints. For each candidate $v_{t,i}\in \mathcal{V}t$ generated by the Waypoint Predictor, we select a viewpoint image of the view in that direction and the corresponding description. 
Aligning the active sub-instruction with the present visual observation of the potential navigation waypoints, and select a fine-grained waypoint for local progress. We then prompt the multimodal language model with a query that includes (i) the active sub-instruction $s_k$, (ii) a short description of the candidate viewpoints and the images (iii) the textual description for the history movement. The prompt asks MLLM to judge which viewpoint viewpoint will meaningfully advance the sub-goal described by $s_k$. The agent then selects the top-scoring direction $v_{t}$ as the next waypoint to pursue. The low-level motion command corresponding to $v_t$ is executed, causing the agent to navigate from $x_t$ to the new location $x_{t+1}$ and yielding a new observation $o_{t+1}$. In addition, the MLLM is required to output an estimated distance $d_t$ to the landmarks mentioned in $s_k$; if $d_t$ falls below a predefined threshold, the current sub-goal is considered ready to be inspected.

\begin{table*}[!ht]
\caption{Comparison with supervised and zero-shot methods on validation unseen split of R2R-CE. 
\textbf{Bold} denotes the best performance across all zero-shot methods, while \underline{Underlined} indicates the second-best results across zero-shot methods.}
\vspace{10pt}
\label{tab:r2r-ce}
\centering
\begin{tabular}{lcccccc}
    \hline\hline
    Method & TL & NE↓ & nDTW↑ & OSR↑ & SR↑ & SPL↑ \\ \hline
    \multicolumn{7}{c}{\textbf{    Supervised Learning}} \\ \hline
    CMA \citep{hong2022bridging}          & 11.08 & 6.92 & 50.77 & 45 & 37 & 32.17 \\
    RecBERT \citep{hong2022bridging}      & 11.06 & 5.80 & 54.81 & 57 & 48 & 43.22 \\
    BEVBert \citep{an2023bevbert}      & 13.63 & 5.13 & 61.40 & 64 & 60 & 53.41 \\
    Navid \citep{zhang2024navid} & 7.63 & 5.47 & -- & 49 & 37 & 35.90 \\
    ETPNav \citep{an2024etpnav}       & 11.08 & 5.15 & 61.15 & 58 & 52 & 52.18 \\ \hline

    \multicolumn{7}{c}{\textbf{Zero-Shot}} \\ \hline
    Random              &  8.15 &  8.63 & 34.08 & 12 &  2 &  1.50 \\
    LXMERT \citep{hong2022bridging}      & 15.79 & 10.48 & 18.73 & 22 &  2 &  1.87 \\
    DiscussNav \citep{long2024discuss}  &  6.27 &  7.77 & 42.87 & 15 & 11 & 10.51 \\
    MapGPT-CE \citep{chen2024mapgpt}   & 12.63 & 8.16 & --    & 21 &  7 &  5.04 \\
    NavGPT-CE \citep{zhou2024navgpt} & -- & 8.37 & -- & 27 & 16 & 10.20 \\
    Open-Nav \citep{qiao2025open}    &  7.68 & \underline{6.70} & \underline{45.79} & 23 & 19 & 16.10 \\
    AO-Planner \citep{chen2025affordances} & -- & 6.95 & -- & 38 & 25 & 16.60 \\
    CA-Nav \citep{chen2025canav} & -- & 7.58 & -- & 48 & 25 & 10.80 \\
    SmartWay \citep{shi2025smartwayenhancedwaypointprediction}           & 13.09 & 7.01 & --    & \textbf{51} & \underline{29} & \underline{22.46} \\
    \textbf{Ours} 
                        &  9.18 & \textbf{5.87} & \textbf{57.70} & \underline{39} & \textbf{34} & \textbf{29.12} \\ 
    \hline\hline
\end{tabular}
\end{table*}

\subsection{Look backward}
\label{looking_backward}
As shown in Fig. ~\ref{fig:example} (b), the look backward module provides trajectory-level verification to catch cumulative drift before the agent terminates. After each sub-instruction—or whenever the agent believes the goal is reached—it assembles a compact textual replay of the visited viewpoints: a chronologically ordered list of landmark names, object mentions, and distances traveled. This replay, together with the finished sub-instructions, is fed back to the frozen MLLM with a prompt inspect two questions: (i) whether the current trajectory satisfies each sub-instruction in order and (ii) whether any overlooked landmark or missed turn suggests a correction. After generate the answer about these questions, the agent should invoke invoke one of four meta-abilities: 

\noindent\emph{continue.} If the distances and audit confirm the current sub-goal is satisfied, advance to $s_{k+1}$.

\noindent\emph{stay.} If signals are borderline or uncertain, remain at $x_t$ and re-query the MLLM without changing the trajectory. 

\noindent\emph{backtrack.} If the audit fails, roll back to the last reliable waypoint $x_r$ and truncate the trajectory $\tau\!\leftarrow\!\tau_{0:r}$.

\noindent\emph{look-around.} If uncertainty is high, temporarily visit all candidate neighbor viewpoints $v\!\in\!\mathcal{N}(x_t)$ to collect observations from these neighbor nodes, then return to $x_t$ for re-evaluation.

By closing this audit loop at runtime—without gradient updates or environment-specific heuristics—looking backward markedly reduces premature stops and large navigation errors, ensuring that global intent aligns with the final executed route.

\section{EXPERIMENTS}
\subsection{Experiment Setup}
\paragraph{Dataset}
We conduct our framework on two standard benchmarks for vision-and-language navigation in continuous environments: \citep{ku2020room} and RxR-CE \citep{krantz2020beyond}. R2R-CE extends the Room-to-Room dataset \citep{anderson2018vision} to continuous settings based on the Habitat simulator \citep{savva2019habitat}. Compared to R2R-CE, RxR-CE introduces longer instructions, longer paths, stricter physical restrictions, and greater risk of getting stuck, making it significantly more challenging. Following the setting in previous work \citep{qiao2025open}, we use the same 100 selected episodes from the val-unseen validation splits of R2R-CE, and randomly sampled 100 episodes from the English val-unseen split of RxR-CE, to balance coverage and API efficiency. For each episode, the agent receives a natural language instruction and must reach the corresponding goal in the Habitat simulator, with the gpt-5-2025-08-07 API serving as the core of the multimodal planner.

\begin{table}[!t]
\caption{Comparison with supervised and zero-shot methods on validation unseen split of RxR-CE.}
\vspace{6pt}
\label{tab:rxr-ce}
\centering
\resizebox{\columnwidth}{!}{
\begin{tabular}{lcccc}
    \hline\hline
    Method & NE↓ & nDTW↑ & SR↑ & SPL↑ \\ \hline
    \multicolumn{5}{c}{\textbf{Supervised Learning}} \\ \hline
    Seq2Seq \cite{krantz2020beyond}   & 12.10 & 30.8 & 13.9 & 11.9 \\
    DC-VLN \cite{hong2022bridging}    &  8.98 & 46.7 & 27.1 & 22.7 \\
    Navid \cite{zhang2024navid}     &  8.41 &  --  & 23.8 & 21.2 \\
    ETPNav \cite{an2024etpnav}    &  5.64 & 61.9 & 54.8 & 44.9 \\ \hline
    
    \multicolumn{5}{c}{\textbf{Zero-Shot}} \\ \hline
    CLIP-Nav \citep{dorbala2022clipnavusingclipzeroshot} & -- & -- & 9.8 & 3.2 \\
    A2Nav \cite{chen20232}     &   --  &  --  & 16.8 &  6.3 \\
    AO-Planner \cite{chen2025affordances}& 10.75 & 33.1 & \textbf{22.4} & 15.1 \\
    CA-Nav \cite{chen2025canav}    & 10.37 & 13.5 & 19.0 &  6.0 \\
    \textbf{Ours}     & \textbf{9.21} & \textbf{45.7} & 22.0 & \textbf{16.1} \\ 
    \hline\hline
\end{tabular}}
\end{table}

\paragraph{Evaluation Metrics}
We utilize the pre-defined evaluation metrics of the VLN task. (1) Trajectory Length (TL), the total distance traveled by the agent; (2) Navigation Error (NE), the shortest geodesic distance between the final position and the goal; (3) Normalized Dynamic Time Warping (nDTW), trajectory similarity by aligning the executed and reference paths; (4) Success Rate (SR), the percentage of episodes where the agent stops within 3 meters of the goal; (5) Object Success Rate (OSR), Success rate of reaching the goal along the trajectory; (6) Success weighted by Path Length (SPL), weighting successful trajectories according to the ratio of the shortest-path distance to the actual path length.

\begin{table}[!t]
\caption{Ablation study on the validation unseen split of R2R-CE. We compare the full Hierarchical Global–Local Planner with ablated variants that disable global reasoning modules.}
\scriptsize
\vspace{6pt}
\label{tab:r2r-ce-ablation}
\centering
\resizebox{\columnwidth}{!}{
\begin{tabular}{l|ccc|ccc}
    \hline\hline
    Method & Step1 & Step2 & Step3 & NE↓ & nDTW↑ & SR↑ \\ \hline
    Ours       & \ding{51} & \ding{51} & \ding{51} & \textbf{5.87} & \textbf{57.70} & \textbf{34} \\
    Ours (only local view)     &  -- & \ding{51} & -- & 6.55 & 54.52 & 20 \\
    Ours (w/o Look Backward)  & \ding{51} & \ding{51} & -- & 6.22 & 55.85 & 28 \\
    \hline\hline
\end{tabular}}
\end{table}

\subsection{Experimental Results}

Table~\ref{tab:r2r-ce} shows a comprehensive performance comparison of both supervised learning and zero-shot methods on the val-unseen split of R2R-CE.
Supervised models, trained with extensive domain-specific data, naturally obtain higher success rates (up to 52--60\%) and maintain strong results across all metrics, highlighting the advantage of task-tailored learning in controlled settings. In contrast, zero-shot agents are deployed without task-specific fine-tuning and generally achieve lower SPL and SR, reflecting the inherent difficulty of transferring general reasoning abilities to embodied navigation. 
Within this challenging zero-shot regime, our Three-Step Nav achieves the best overall performance, reaching an SR of 34\%, an NE of 5.87, and an SPL of 29.12\%. Compared with the strongest prior zero-shot SOTA methods, this corresponds to an improvement of 5\% in SR and 6.66\% in SPL, meanwhile we gain a relatively reduction of 12.4\% in NE. These improvements indicate that our Hierarchical Global-Local Planner framework enables more decision-making and avoids unnecessary navigation error, resulting in stronger goal-reaching capability and trajectory efficiency. Overall, the results demonstrate that incorporating structured global–local reasoning into an MLLM-based planner significantly boosts zero-shot performance and narrows the gap with fully supervised navigation systems.

Table~\ref{tab:rxr-ce} reports the results on the RxR-CE dataset. 
Supervised approaches such as ETPNav, which are trained with extensive task-specific data, achieve strong performance (e.g., SR of 54.8\% and SPL of 44.9), but require large-scale supervision that limits generalizability. 
In contrast, zero-shot models operate without fine-tuning and generally lag behind in SPL and SR. 
Within this challenging regime, our Three-Step Nav achieves competitive results, with an NE of 9.21, an nDTW of 45.7, an SR of 22.0, and an SPL of 16.1. 
Compared with the strongest zero-shot baselines, our method improves nDTW by 12.6\%, and relatively reduces NE by 11.2\%. These results highlight that the integration of global–local reasoning and trajectory-level auditing enables robust progress estimation in long-horizon navigation, even under the more demanding RxR-CE benchmark.

\begin{table}[!t]
\caption{Comparison of different multimodal LLM experts on the validation unseen split of R2R-CE. We report results using our full Hierarchical Global–Local Planner.}
\scriptsize
\vspace{6pt}
\label{tab:r2r-ce-mlmms}
\centering
\begin{tabular}{l|ccc}
    \hline\hline
    MLLM Expert & NE↓ & nDTW↑ & SR↑ \\ \hline
    GPT-5   & \textbf{5.87} & \textbf{57.70} & \textbf{34} \\
    GPT-4o  & 6.70 & 56.42 & 28 \\
    GPT-4v  & 6.94 & 54.10 & 26 \\
    \hline\hline
\end{tabular}
\end{table}

\subsection{Ablation Study}
Table~\ref{tab:r2r-ce-ablation} reports the ablation results on R2R-CE, highlighting the contributions of different modules in our hierarchical design. 
The full model with all three steps enabled achieves the strongest performance (NE 5.87, nDTW 57.70, SR 34\%). 
When only the local reasoning module (\emph{look now}) is active, performance drops substantially (SR 20\%, NE 6.55), showing that local chain-of-thought reasoning alone cannot maintain long-range consistency. 
Enabling global forward planning but removing the backward verification improves results to SR 28\% and nDTW 55.85, yet still lags behind the full framework. 
These comparisons demonstrate that both global planning (\emph{look forward}) and trajectory-level auditing (\emph{look backward}) are indispensable: the forward step provides high-level guidance, while the backward check prevents drift and premature stops. 
Together, they enable our Three-Step Nav to achieve robust zero-shot navigation beyond local-only baselines.

To further understand the role of the underlying MLLM expert, Table~\ref{tab:r2r-ce-mlmms} compares GPT-5, GPT-4o, and GPT-4V within our full Three-Step Nav framework. Among the three tested models, GPT-5 delivers the strongest results (NE 5.87, nDTW 57.70, SR 34\%), demonstrating its superior capability for long-horizon spatial reasoning. 
Using GPT-4o still provides competitive performance, but with a noticeable drop in SR (28\%) and slightly higher navigation error (NE 6.70), while GPT-4v lags further behind with SR 26\% and NE 6.94. 
These comparisons suggest that although our planner consistently benefits from structured global–local reasoning regardless of the underlying expert, more advanced MLLMs such as GPT-5 yield significant gains in both success rate and trajectory fidelity. 
This highlights the importance of model quality when scaling zero-shot navigation frameworks to more complex environments.

\section{CONCLUSIONS}
\paragraph{Summary} In this work, we introduced Three-Step Nav, a hierarchical global–local planner that enables zero-shot VLN in continuous environments without any task-specific training. Our three-stage approach – look forward, look now, and look backward – allows a frozen multimodal LLM to maintain long-horizon context, make fine-grained decisions, and self-correct by backtracking. This framework is lightweight and modular, easily integrating into existing navigation pipelines. Three-Step Nav set a new state-of-the-art among zero-shot methods on R2R-CE, reducing navigation error by about 15\% and substantially improving success rate and path efficiency (SPL) compared to previous approaches. It also achieved strong performance on the challenging RxR-CE benchmark, demonstrating competitive results close to fully supervised policies. These gains validate that coupling global route planning with trajectory-level verification can significantly mitigate distraction and drift in long-horizon navigation.

\paragraph{Method Assumptions and Limitations} While effective, our approach assumes access to a powerful multimodal language model and a relatively static environment. The agent’s reasoning heavily relies on the pre-trained LLM (e.g. GPT-5), which introduces computational cost and potential errors if the model misinterprets visual cues or instructions. Also, our evaluations are in simulated indoor environments – transferring to real robots will require handling sensor noise, moving obstacles, and other real-world complexities. The Three-Step Nav framework does not learn from feedback over time, so extremely long or ambiguous instructions can still pose challenges. In future work, we plan to refine the system’s robustness by incorporating learning-based adaptation or fine-tuning smaller navigation-specific models. Additionally, optimizing the prompting strategy and LLM integration could reduce latency, making the system more feasible for real-time robotic deployment. Validating our planner on physical robots in real homes or public spaces is an important next step toward closing the sim-to-real gap.

\paragraph{Societal Impact} Improving Vision-and-Language Navigation has positive implications for assistive robotics and autonomous agents in society. A robust zero-shot navigation agent could assist visually impaired users in unfamiliar indoor spaces or enable home robots to follow complex spoken instructions, enhancing accessibility and convenience. Our method’s ability to self-audit and correct mistakes is particularly valuable for safety-critical applications like search-and-rescue, where backtracking from a wrong turn can prevent accidents. We are mindful that any autonomous navigation system must be thoroughly tested to avoid hazards – for instance, a navigation error in a real home could lead to collisions. By reducing drift and improving reliability, Three-Step Nav contributes to safer deployment of embodied AI. Continued research should examine ethical considerations (such as minimizing biases in language understanding) and include stakeholders in testing to ensure these systems benefit users of diverse backgrounds. Overall, our hierarchical planner represents a step toward more trustworthy and general-purpose embodied agents, bridging the gap between simulation benchmarks and real-world navigation tasks.

\acknowledgments{This work was supported by the National Eye Institute (grant R61EY037527) and the National Science Foundation (award 2318101). The authors affirm that the views expressed herein are solely their own, and do not represent the views of the United States government or any agency thereof.}

\bibliography{aistats}

\section*{Checklist}

\begin{enumerate}

  \item For all models and algorithms presented, check if you include:
  \begin{enumerate}
    \item A clear description of the mathematical setting, assumptions, algorithm, and/or model. [Yes]
    \item An analysis of the properties and complexity (time, space, sample size) of any algorithm. [Yes]
    \item (Optional) Anonymized source code, with specification of all dependencies, including external libraries. [Yes]
  \end{enumerate}

  \item For any theoretical claim, check if you include:
  \begin{enumerate}
    \item Statements of the full set of assumptions of all theoretical results. [Not Applicable]
    \item Complete proofs of all theoretical results. [Not Applicable]
    \item Clear explanations of any assumptions. [Not Applicable]     
  \end{enumerate}

  \item For all figures and tables that present empirical results, check if you include:
  \begin{enumerate}
    \item The code, data, and instructions needed to reproduce the main experimental results (either in the supplemental material or as a URL). [Yes]
    \item All the training details (e.g., data splits, hyperparameters, how they were chosen). [Not Applicable]
    \item A clear definition of the specific measure or statistics and error bars (e.g., with respect to the random seed after running experiments multiple times). [Yes]
    \item A description of the computing infrastructure used. (e.g., type of GPUs, internal cluster, or cloud provider). [Yes]
  \end{enumerate}

  \item If you are using existing assets (e.g., code, data, models) or curating/releasing new assets, check if you include:
  \begin{enumerate}
    \item Citations of the creator If your work uses existing assets. [Yes]
    \item The license information of the assets, if applicable. [Yes]
    \item New assets either in the supplemental material or as a URL, if applicable. [Yes]
    \item Information about consent from data providers/curators. [Not Applicable]
    \item Discussion of sensible content if applicable, e.g., personally identifiable information or offensive content. [Not Applicable]
  \end{enumerate}

  \item If you used crowdsourcing or conducted research with human subjects, check if you include:
  \begin{enumerate}
    \item The full text of instructions given to participants and screenshots. [Not Applicable]
    \item Descriptions of potential participant risks, with links to Institutional Review Board (IRB) approvals if applicable. [Not Applicable]
    \item The estimated hourly wage paid to participants and the total amount spent on participant compensation. [Not Applicable]
  \end{enumerate}

\end{enumerate}

\clearpage
\appendix
\onecolumn
\aistatstitle{Implementation Details}

\section{LOOK FORWARD MODULE}


The \emph{look forward} module serves as a global parser for the navigation
instruction.  It takes the entire instruction as input and decomposes it
into a sequence of atomic actions, ensuring that all actions and
landmark descriptions are captured.  Each sub‑instruction produced by
the module must contain at least one action (such as ``go through the
door'' or ``turn left''), and, when mentioned, it must also include
specific landmarks that ground the action in the environment.  This
global view prevents the omission of actions and ensures the integrity
of the instruction decomposition.

\begin{figure}[!ht] 
    \centering
    \includegraphics[width=\textwidth]{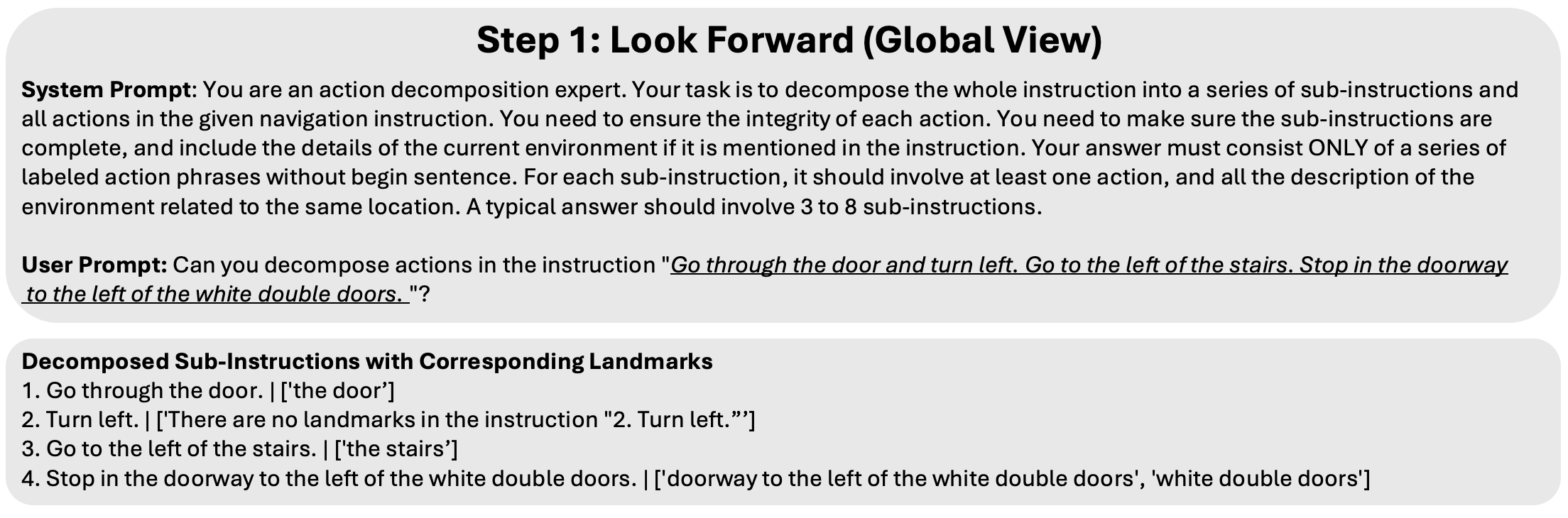}
    \caption{We designed the sub-instruction separation prompts for the \emph{look forward} module. }
    \label{fig:step1}
\end{figure}

As illustrated in Figure~\ref{fig:step1}, the system first provides
an internal \emph{system prompt} describing how to decompose the input
instruction: sub‑instructions should be complete sentences, each
covering a single action and any referenced landmarks.  The \emph{user prompt}
supplies a navigation instruction, for example: ``Go through the door
and turn left.  Go to the left of the stairs.  Stop in the doorway to
the left of the white double doors.''  The module then outputs a list
of sub‑instructions paired with their corresponding landmarks. These sub‑instructions ensure that each action is explicit and that any
environmental details are retained.  A typical output contains
approximately three to eight sub‑instructions, depending on the
complexity of the original instruction, providing a structured and
human‑readable sequence for downstream processing.

\section{LOOK NOW MODULE}
The \emph{look now} module provides a fine-grained, step-wise navigation process conditioned on the current visual observation.
At each step, the system integrates the parsed sub-instruction from the \emph{look forward} stage with the accumulated navigation history and the scene observations around the agent. 
It constructs a detailed system prompt that enumerates the visible candidate viewpoints, their semantic and geometric properties, and the agent’s current spatial context. 
The multimodal large language model (MLLM) is then prompted to reason about which direction best aligns with the current sub-instruction, producing a structured output that contains: (1) a chain-of-thought justification (“Thought”), (2) the estimated distance toward the predicted direction, (3) the predicted next viewpoint, and (4) a binary completion estimation indicating whether the current sub-instruction is complete.

\begin{figure}[!ht] 
    \centering
    \includegraphics[width=\textwidth]{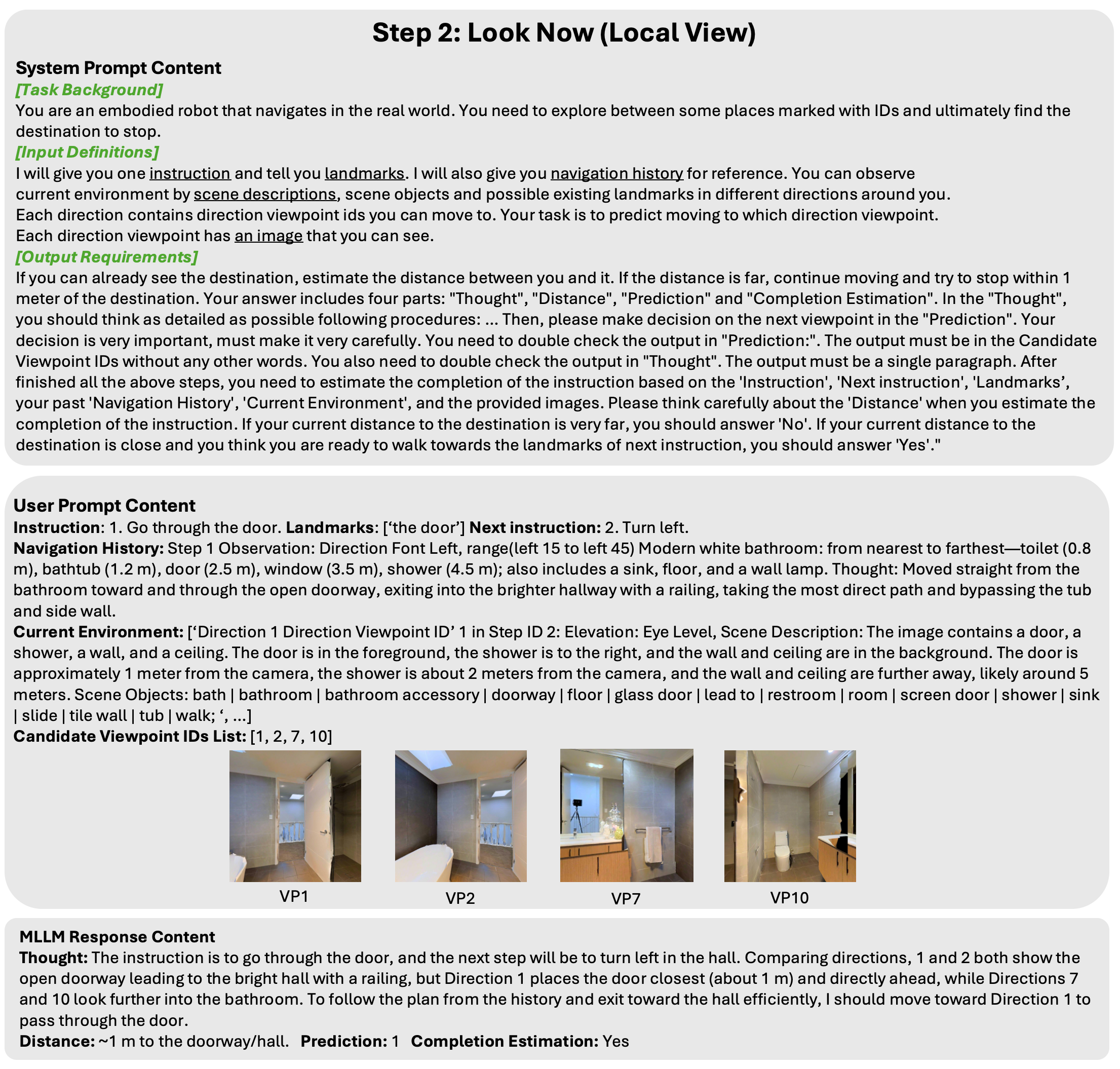}
    \caption{Task prompts for the local-level \emph{look now} module. }
    \label{fig:step2}
\end{figure}

As illustrated in Figure~\ref{fig:step2}, the MLLM compares candidate viewpoints by analyzing spatial cues (e.g., door positions, hallway direction) and aligns them with linguistic landmarks in the instruction.
In this example, the model determines that the next action “turn left” should follow after moving through the door.
It selects the viewpoint corresponding to the open hallway (Direction 1) as the most consistent continuation and estimates that the local goal is reached.

\section{LOOK BACKWARD MODULE}

\begin{figure}[!ht]
    \centering
    \includegraphics[width=0.8\textwidth]{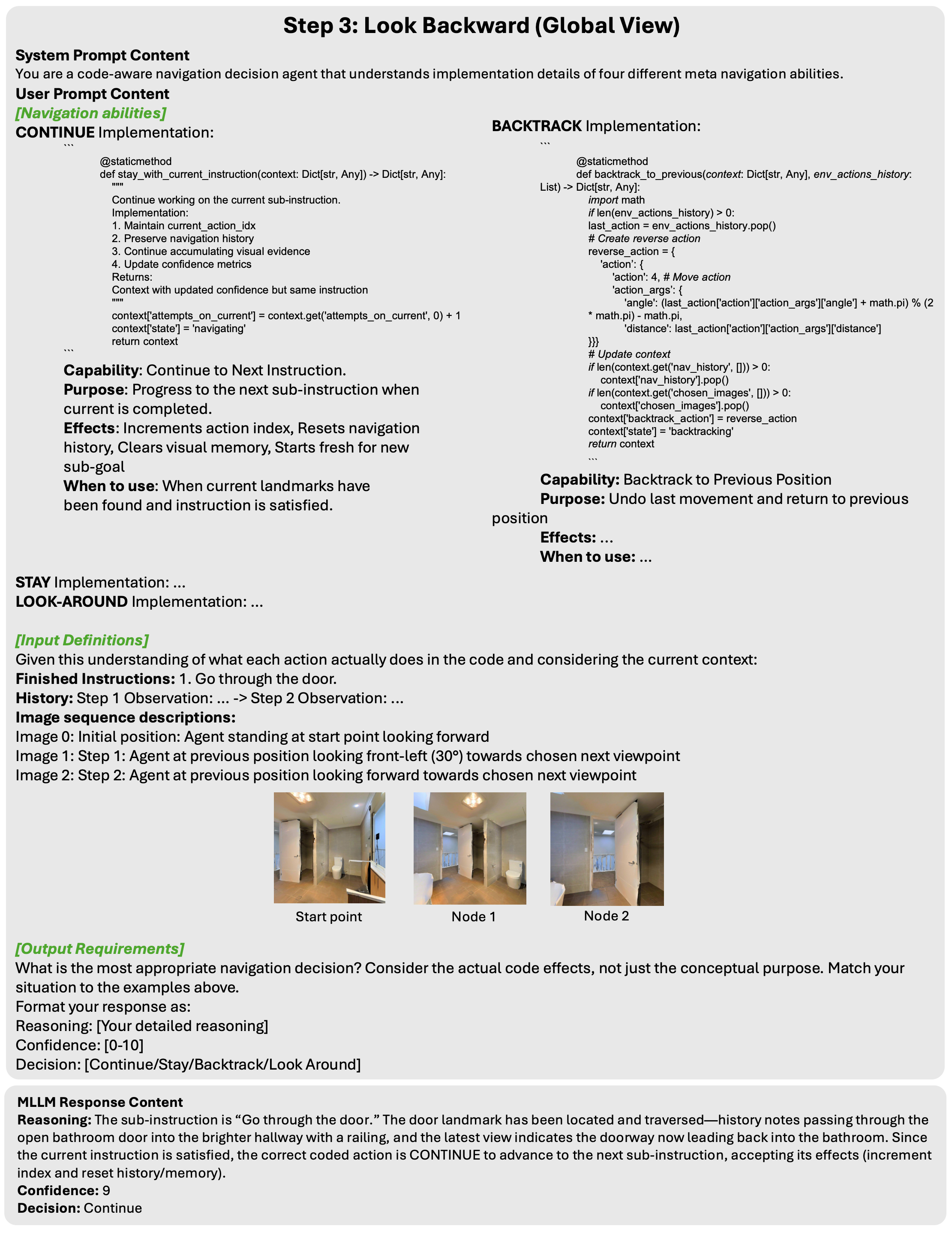}
    \caption{
    An example of the complete prompting process in the \emph{look backward} module.
    The system integrates both visual and code-level contexts to decide how to proceed after executing a sub-instruction.
    The prompt includes code definitions for four meta-navigation abilities—\emph{Continue}, \emph{Stay}, \emph{Look-Around}, and \emph{Backtrack}—together with sequential observations of the agent’s trajectory.
    Here, the MLLM reasons that the sub-instruction “go through the door” has been fulfilled and correctly selects \emph{Continue}, indicating readiness to progress to the next navigation goal.
    }
    \label{fig:step3}
\end{figure}

When the progress estimation result from \emph{look now} module shows 'yes', ThreeStep Nav will start the \emph{look backward} module.
As illustrated in Figure~\ref{fig:step3}, the \emph{look backward} module acts as a global auditing mechanism that evaluates whether the executed actions have satisfied the current sub-instruction and determines how to transition to the next one. 
While the \emph{look forward} and \emph{look now} modules operate on linguistic parsing and local visual reasoning, respectively, the \emph{look backward} stage explicitly integrates code-level semantics and navigation control logic.
The model is prompted with both the high-level navigation history and the code implementation snippets that define meta-navigation behaviors, such as \emph{Continue}, \emph{Stay}, \emph{Look-Around}, and \emph{Backtrack}. 
By understanding the effects and intended purposes of these functions, the system can reason about which control operation best fits the current context.

\vfill


\end{document}